\documentclass{article}
\usepackage[preprint]{neurips_2025}
\usepackage{graphicx}
\usepackage[utf8]{inputenc} 
\usepackage[T1]{fontenc}    
\usepackage{hyperref}       
\usepackage{url}            
\usepackage{booktabs}       
\usepackage{amsfonts}       
\usepackage{amsmath}
\usepackage{nicefrac}       
\usepackage{microtype}      
\usepackage{xcolor} 
\usepackage{cleveref}
\usepackage{algorithm}
\usepackage{algpseudocode}
\usepackage{wrapfig}

\title{Controllable diffusion-based generation for multi-channel biological data}

\author{%
  Haoran Zhang \\
  University of Texas at Austin\\
  Austin, TX 78712 \\
  \texttt{hz6453@utexas.edu} \\
  \And
  Mingyuan Zhou \\
  University of Texas at Austin \\
  Austin, TX 78712 \\
  \texttt{mingyuan.zhou@mccombs.utexas.edu} \\
  \And
  Wesley Tansey \\
  Memorial Sloan Kettering Cancer Center \\
  New York, NY 10065 \\
  \texttt{tanseyw@mskcc.org } \\
}

\begin{document}

\maketitle

\begin{abstract}
    Spatial profiling technologies in biology, such as imaging mass cytometry (IMC) and spatial transcriptomics (ST), generate high-dimensional, multi-channel data with strong spatial alignment and complex inter-channel relationships. Generative modeling of such data requires jointly capturing intra- and inter-channel structure, while also generalizing across arbitrary combinations of observed and missing channels for practical application. Existing diffusion-based models generally assume low-dimensional inputs (e.g., RGB images) and rely on simple conditioning mechanisms that break spatial correspondence and ignore inter-channel dependencies. This work proposes a unified diffusion framework for controllable generation over structured and spatial biological data. Our model contains two key innovations: (1) a hierarchical feature injection mechanism that enables multi-resolution conditioning on spatially aligned channels, and (2) a combination of latent-space and output-space channel-wise attention to capture inter-channel relationships. To support flexible conditioning and generalization to arbitrary subsets of observed channels, we train the model using a random masking strategy, enabling it to reconstruct missing channels from any combination of inputs. We demonstrate state-of-the-art performance across both spatial and non-spatial prediction tasks, including protein imputation in IMC and gene-to-protein prediction in single-cell datasets, and show strong generalization to unseen conditional configurations.
\end{abstract}

\section{Introduction}

Recent advances in generative models enable structured generation, prediction, and imputation across various data domains \cite{Rombach:2022:LDM,Zhang:2023:ControlNet}. Specifically, diffusion models have shown a remarkable capacity for generating high-fidelity samples with corrupted data or conditionals like text prompts, especially in natural images and language generation tasks. However, the domain of biological profiling data, including imaging mass cytometry (IMC) \citep{chang:etal:2017:imc-review}, spatial transcriptomics (ST) \citep{moses:pachter:2022:museum-of-st}, and other biological profiling technologies, poses unique structural challenges that remain underexplored.

A key characteristic of biological profiling data is its many-channel nature, where each dataset comprises a substantial number of channels ($n \geq 30$ for proteomics technologies and $n \geq 1000$ for transcriptomics technologies). For instance, in spatial profiling data, each channel designates a specific molecule of interest (e.g., proteins and genes), and each pixel (or cell) represents a spatially co-registered vector of biologically distinct signals. Unlike RGB images with fixed and highly correlated channels, biological channels often have complex and variable inter-channel correlations. More importantly, in practice, many channels may be unobserved due to technical limitations, acquisition constraints, or experimental design. This constraint forces biologists to decide which signals are most important to measure before even seeing the data, limiting the scope and breadth of insights gained from an experiment. A generative framework that can handle high-dimensional biological profiling data, maintain spatial alignment, accommodate missing channels, and model diverse inter-channel relationships could potentially fill in missing channels. Such a method would enhance the scope of biological studies and potentially lead to more scientific discoveries. 

This problem is more complex than conditional natural image generation. The conditioning inputs in this task are not abstract prompts (e.g., text or class labels), but observed channels of the same sample. These conditions are spatially aligned with the target, and such spatial alignment should be preserved throughout the generation process. Naive conditioning mechanisms, such as global embeddings and flat concatenation, break spatial correspondence, making them incompatible with biological profiling tasks. On the other hand, the number of channels is large and their interactions are highly context-dependent. Some channels co-localize only within specific spatial niches or cell types; others may be mutually exclusive. Therefore, modeling such sparse, nonlinear, and asymmetric dependencies requires a generative framework that maintains spatial alignment and adapts to diverse and context-dependent channel semantics in biological data. 

Existing data imputation and colorization methods, including score-based~\citep{Song:2019:score} and conditional diffusion models~\citep{Saharia:2022:Palette}, offer powerful tools for spatially constrained image generation. However, these methods are tailored to low-dimensional RGB images and use generic conditioning strategies that ignore channel semantics. ControlNet~\citep{Zhang:2023:ControlNet} and BrushNet~\citep{Ju:2024:BrushNet} introduce multiscale spatial conditioning mechanisms but remain limited to low-dimensional RGB-like inputs and rely on pre-trained diffusion models, which do not model inter-channel relationships in a principled way. As a result, such methods fail to generalize to the high-dimensional setting of biological data, where conditional generalization and principled modeling of inter-channel relationships are important.
\begin{figure} 
\label{fig:visual_comparison}
\includegraphics[width=1\linewidth]{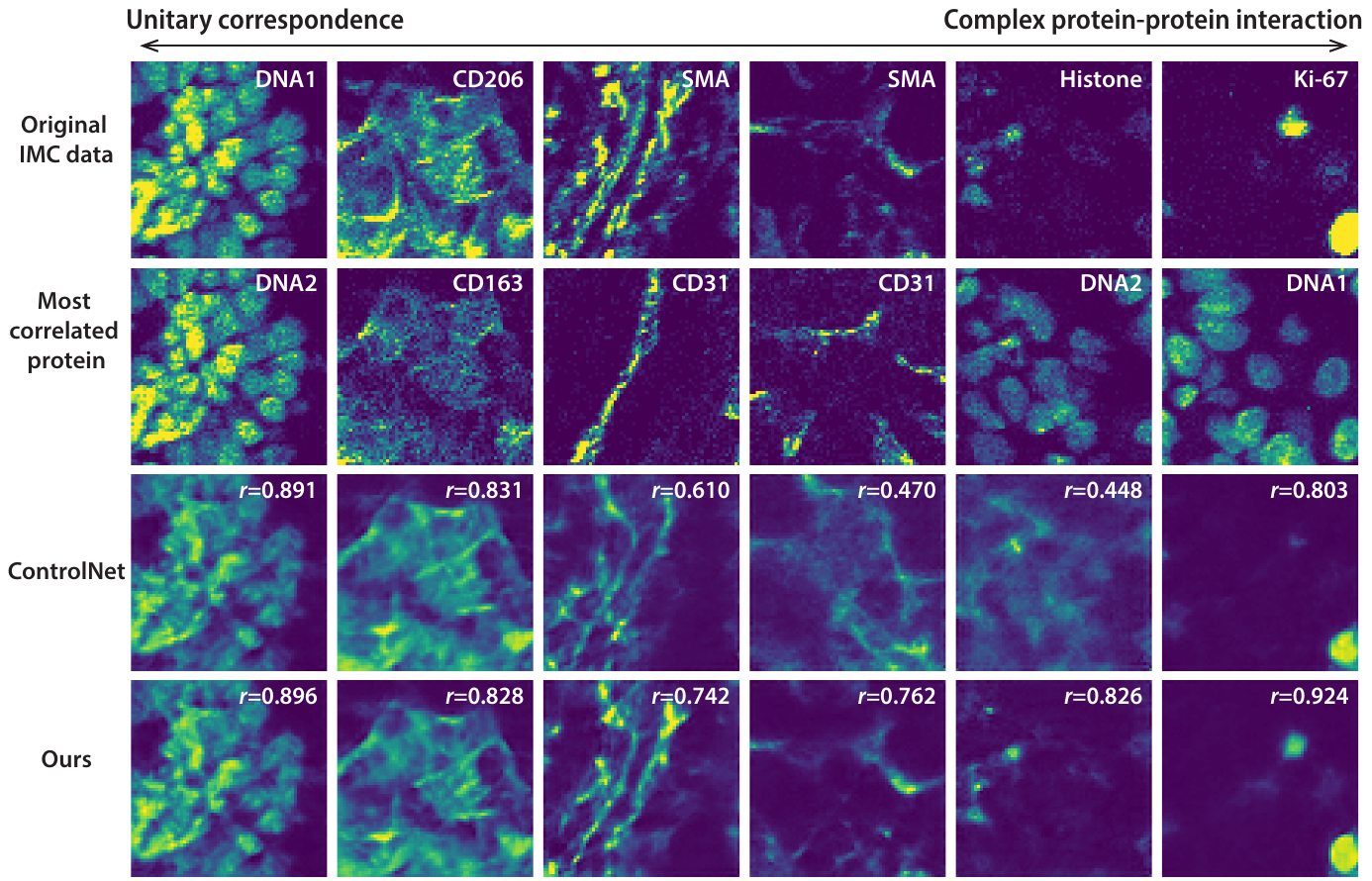} 
\caption{Visual comparison of samples generated by our model with samples generated by ControlNet-style counterpart. Without the channel attention modules and attention injection mechanisms, ControlNet-style generative models can handle simple channels with unitary inter-channel correlation, like DNA1 (the right-most column), but easily fail in the ones like tumor markers that have more complex protein-protein interaction like Ki-67 and $\alpha$SMA (columns 3-6).} 
\end{figure} 

In this work, we introduce a diffusion-based generative model for controllable multi-channel data generation with strong spatial alignment but heterogeneous channel semantics. We demonstrate its performance and generalizability in tasks of generating biological profiling data with subsets of observable channels. Our method handles arbitrary combinations of observed and missing channels, maintains spatial alignment, and captures complex inter-channel dependencies. To achieve this, we propose three key components: (1) a hierarchical feature injection mechanism that conditions the generation of multi-resolution spatial features on the observed channels, preserving alignment and enabling fine-grained control, (2) a combination of latent- and output-space channel-wise attention to explicitly model semantic relationships between channels during generation, and (3) a random masking training strategy that amortizes over the conditional space and enables dynamic test time conditioning. Together, these components allow our model to generate multi-channel biological profiling data with high fidelity across resolution scales within a unified diffusion framework. Our model achieves state-of-the-art results in both spatial and non-spatial settings and supports test-time controllability over arbitrary conditioning on subsets of channels.

\section{Background}
\subsection{Diffusion models}
Diffusion models are a class of generative models that approximates the data distribution by modeling the gradient of its log-density (score function) \citep{Song:2019:score}. This is achieved by coupling a forward process that progressively perturbs the data with a reverse process that learns to undo this perturbation.
The forward process transforms data $\mathbf{x}_0$ into a noise distribution $\mathbf{x}_T$ over $T$ timesteps by injecting noise from a known distribution like a Gaussian \citep{Sohl:2015:DPM},
\begin{equation}
q(\mathbf{x}_t \mid \mathbf{x}_0) = \mathcal{N}(\mathbf{x}_t \mid \sqrt{\alpha_t} \mathbf{x}_0, (1 - \alpha_t) \mathbf{I}),
\end{equation}
where $\alpha_t$ is a noise schedule parameterizing the variance at each timestep $t$. As $t$ approaches $T$, the data distribution converges to a simple prior, such as a standard Gaussian.
The reverse process reconstructs data from noise by learning to reverse the corruption introduced during the forward process. This is achieved through a neural network $\epsilon_\theta(x_t, t)$, trained to predict the noise $\epsilon$ added at each timestep \citep{Ho:2020:DDPM}. The denoising process is guided by the objective:
\begin{equation}
\mathcal{L}_{\text{DDPM}} = \mathbb{E}_{\mathbf{x}_0, \mathbf{\epsilon}, t} \left[ \| \mathbf{\epsilon} - \mathbf{\epsilon}_\theta(\mathbf{x}_t, t) \|^2 \right],
\end{equation}
where $\mathbf{\epsilon} \sim \mathcal{N}(0, \mathbf{I})$ is the Gaussian noise injected by the forward process. By minimizing this objective, the model learns to reconstruct $x_0$ from the noisy $x_t$ at any timestep $t$. This denoising approach is deeply connected to score matching. Specifically, the noise prediction $\epsilon_\theta(x_t, t)$ can be used to compute the gradient of the log-density (score function) as:
\begin{equation}
\nabla_{\mathbf{x}_t} \log q(\mathbf{x}_t) \propto -\frac{\mathbf{\epsilon}_\theta(\mathbf{x}_t, t)}{\sqrt{1 - \alpha_t}}
\end{equation}
Thus, accurately predicting the noise is equivalent to estimating the score function $\nabla_{\mathbf{x}_t}\log q(\mathbf{x}_t)$, which guides the reverse process. This insight bridges denoising and score matching, making the reverse process a refinement procedure that progressively moves noisy samples back to the data manifold \cite{Song:2019:score}.


\subsection{Structured multi-channel imputation}
Image imputation is a classical problem in computer vision and generative modeling, where the objective is to recover missing or corrupted parts of an image given the observed context. Formally, given an observation $c$, classical imputation methods aim to estimate the full image $x$ by modeling the conditional distribution $p(x \mid c)$. This problem has been extensively studied in the context of RGB images, where $x, v \in \mathbb{R}^{3 \times H \times W}$, and inpainting primarily relies on spatial continuity within the image domain \cite{chan2001nontexture}.

We generalize this problem to the setting of structured multi-channel data, where each channel corresponds to a semantically distinct measurement—such as a spectral band, a protein marker, or a gene expression. Let $c \in \mathbb{R}^{C_o \times H \times W}$ denote the observed data, and let $x \in \mathbb{R}^{C \times H \times W}$ denote the full data, where $C = C_o + C_m \geq 1$, $C_o \geq 1$, and $C_m \geq 0$ denote the number of total, observed, and missing channels respectively. Note that when $C_m =0$ and $C_o = C = 3$, this formulation reduces to the classical RGB image imputation problem ($C=1$ for the grayscale case), and when $C_m = 3$ and $C_o = 1$, it reduces to the classical RGB image colorization problem. 

This general formulation applies across a broad range of problems at different resolution. When $H = W = 1$, the input reduces to a high-dimensional vector, making this formulation applicable to non-spatial data imputation such as single-cell data, where one may predict single cell protein expression from single-cell RNA (scRNA) sequencing data. When $H, W > 1$, the formulation supports spatially structured imaging tasks such as IMC channel prediction, where local morphology and global tissue organization both contribute to the reconstruction target.

\section{Methods}
We propose a diffusion-based generative framework for multi-channel biological data conditioned on arbitrary subsets of observed channels. Let $x \in \mathbb{R}^{C\times H\times W}$ be the data with a full set of channels and $c \in \mathbb{R}^{C_o\times H\times W}$ the observed subset. Our goal is to learn the conditional distribution $p(x|c)$, where $x$ and $c$ are spatially aligned.

\begin{figure} 
\includegraphics[width=1\linewidth]{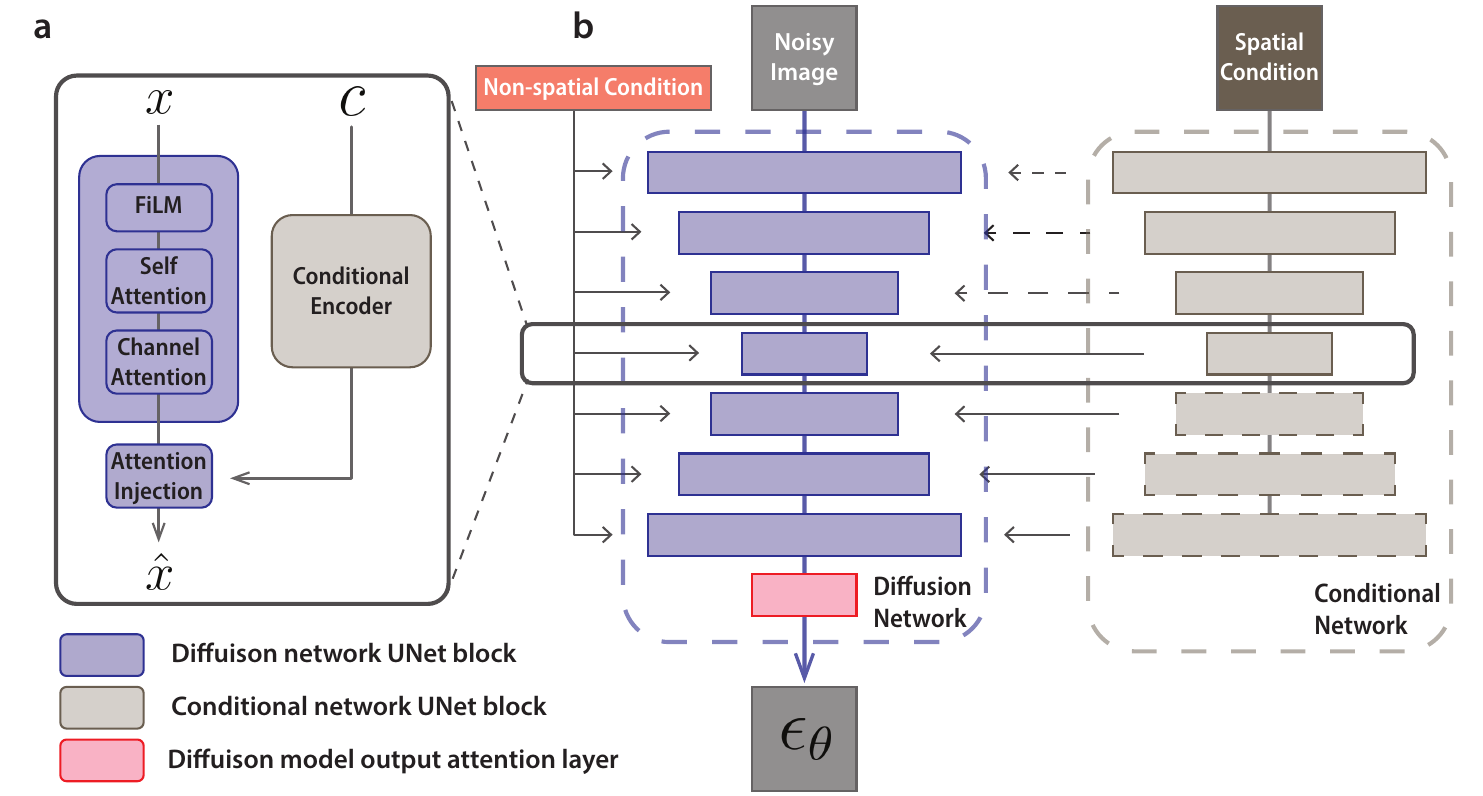} 
\caption{\label{fig:model} Overview of the proposed diffusion model for multi-channel data generation. (a) Custom channel-wise attention: UNet Block with channel attention module that models the inter-channel relationships. (b) Hierarchical feature injection: a parallel conditional network encodes the spatial condition, producing features at different resolution that are injected into the corresponding UNet block in the diffusion network.} 
\end{figure} 

\subsection{Spatially Aligned Feature Injection for Structured Conditioning} 
We first introduce a hierarchical feature injection mechanism that conditions $x$ on multi-resolution representations of the observed $c$. We employ two jointly trained parallel networks: a diffusion network that processes the noisy target $x_t$, and a contextual network that processes the observed spatial condition $c$. At each resolution level $\ell$, the diffusion network encodes $x_t$ to $D_\ell(x_t)$, and the contextual network gives the contextual feature map $E_\ell(c)$, which is injected into the corresponding layer of the diffusion network (\cref{fig:model}b). These contextual features are injected through a channel-wise gating module: 
\begin{equation}
\begin{aligned}
z_\ell  &= D_\ell(x_t) + \text{SE}(E_\ell(c))\\
\end{aligned}
\end{equation} 
allowing for the flexible injection of spatial conditions. The $\text{SE}(\cdot)$ is the Squeeze-and-Excitation block that serves as soft channel attention (see \cref{sec:channel_attention} for detail) and selectively injects the condition while the spatial alignment is preserved. This design allows the model to condition not just on a fixed global representation of $c$ but on a series of contextual features $\{E_\ell(c)\}_{\ell=1}^L$ that vary across resolution. This setup reflects the intuition that certain patterns in $x$, e.g. global motifs versus local structures, may depend on different aspects of $c$. Early encoder layers focus more on the local structures, while late layers draw high-level, global structures.  

\subsection{Training with Random Channel Masking} 
Different experiments may measure different subsets of possible channels. Rather than training separate models to predict individual channels, our goal is to produce full multi-channel outputs (i.e. the complete protein panel). Constructing our model so that it can condition on any subset enables us to train across multiple datasets and flexibly apply our model on each. To create a unified model, we propose a random channel masking strategy during training. In each iteration, we randomly sample a subset of channels $S_o \subset \{1, \dots, C\}$ as observed, and mask out the remaining $S_m = \{1, \dots, C\} \backslash S_o$. The masked subset $x_{S_m}$ is zeroed out in spatial condition $c$, and the model reconstructs the full panel $x \in \mathbb{R}^{C \times H\times W}$. 

\begin{wrapfigure}{R}{0.55\textwidth}
\vspace{-25pt}
    \begin{minipage}{0.54\textwidth}
      \begin{algorithm}[H]
      \caption{Random Channel Masking}
      \begin{algorithmic}[1]
\Require Full data $x \in \mathbb{R}^{C \times H \times W}$, masking prob. $p$
\For{each training iteration}
    \State Sample observed set $S_o \subset \{1, \dots, C\}$,\newline where $I_{i\in S_o} \sim Bern(p)$
    \State Construct conditon $c$ such that
            \[
            c_i = 
            \begin{cases}
            x_i, & \text{if } i \in S_o \\
            0,    & \text{otherwise}
            \end{cases}
            \]
    \State Diffusion step with target $x$ and condition $c$
\EndFor
\end{algorithmic}
\end{algorithm}
\end{minipage}
\vspace{-14pt}
\end{wrapfigure}

Randomly varying $S_o$ during training encorages the model to learn conditional generation under diverse partial contexts. As a result, the trained model can generalize to arbitrary combinations of observed and missing channels at the test time, including unseen configurations. At the same time, because the model always predicts all channels, it avoids the need for channel-specific heads or separate training for each protein, and enables comprehensive downstream analysis and one-stop training.

This training procedure implicitly defines an amortized conditional learning objective:
\begin{equation} \label{eqn:amortization}
    \mathbb{E}_{c \sim P(c)}\mathbb{E}_{t, x_0, \epsilon}\left[||\epsilon - \epsilon_\theta(x_t, t, c)||^2\right]
\end{equation}
where $p(c), c\in \mathcal{C}$ is the distribution of conditional configuration (i.e. combinations of observable channels) over the conditional space $\mathcal{C}$. Therefore, one can treat the random masking training as an amortized inference over the condition space. Minimizing \cref{eqn:amortization} encourages the model to learn a single amortized estimator $\epsilon_\theta(x_t, t, c)$ such that $\epsilon_\theta(x_t, t, c) \approx \nabla_{x_t}\log p(x_t|c)$ for arbitray conditon $c$. This strategy relies on the assumption that partially observed structured data can provide informative gradients in amortized training. Although not formally addressed, previous work has shown its effectiveness in various domains \cite{gershman2014amortized, marino2018iterative}, and we provide empirical justification in \cref{sec:exp_spatial}.

\subsection{Channel Attention for Structured Feature Modulation} \label{sec:channel_attention}
With random channel masking, our model aims to handle arbitrary conditional configurations, where the observed channels may vary across samples and tasks. Since all condition $c$ are encoded with the same conditional network, the model needs to adaptively select and reweight latent features depending on the conditions. 
To enable the model to dynamically recalibrate features in both latent and data space, 
we introduce a bespoke channel-wise attention module in each UNet block (\cref{fig:model}a).
We consider two designs, each focusing on distinct modeling goals.

The first is a lightweight attention mechanism inspired by the Squeeze-and-Excitation (SE) network \cite{Hu:2018:SENet}. Given a latent feature map $z \in \mathbb{R}^{D \times H \times W}$, we compute
\begin{equation}
\begin{aligned}
\alpha &= \text{GAP}(z)\\
w &= \sigma\left( W_2 \cdot \phi(W_1 \cdot \alpha) \right) \\
z' &= w \cdot z \\
\end{aligned}
\end{equation} 
where $\phi$ is a non-linearity (e.g., ReLU), $\text{GAP}$ stands for global average pooling, and $\sigma$ is the sigmoid function. This approach scales each latent channel by a learned weight conditioned on the global context and enables a per-sample feature reweighting that aligns with the dynamic feature selection requirement for the random channel masking. 

Alternatively, we can perform a full channel-wise self-attention across all channels: 
\begin{equation} 
x_{\text{flat}} \in \mathbb{R}^{D \times N}, \quad N = H \times W, \quad Q = x_{\text{flat}} W_Q, \quad K = x_{\text{flat}}W_K, \quad V = x_{\text{flat}} W_V 
\end{equation} 
\begin{equation} \label{eqn:kqv_ch_attn} 
A = \text{softmax}\left( \frac{Q K^\top}{\sqrt{d}} \right), \quad x_{\text{flat}}' = A V 
\end{equation} 
Compared to SE channel attention, this design is more expressive and can capture higher-order dependencies among channels. This feature makes the transformer-based channel attention particularly useful within the UNet blocks, where the model infers the missing information across latent features.

We introduce an additional channel attention mechanism at the final stage of the model, operating over the semantic output channels. Let the final latent representation before projection be $z \in \mathbb{R}^{D \times H \times W}$, where $D$ is the number of latent channels. The standard diffusion model produces $\hat{y} \in \mathbb{R}^{C \times H \times W}$ with a single output layer. To model cross-channel dependencies, we add an additional layer that computes
\begin{equation}
\hat{y}_{\text{attn}} = y + \text{Conv1}(\text{SE}(y))
\end{equation}
Together, these two attention modules allow the model to modulate features both within the latent space and across semantic output channels, improving performance on tasks involving complex, structured channel relationships.

\section{Related Work}
\paragraph{Image inpainting with conditional diffusion models.}
Recent work has applied diffusion-based generative models to image inpainting and completion tasks \citep{Song:2019:score, Saharia:2022:Palette}. These models primarily work on natural and grayscale images, where the number of channels is limited ($n \leq 3$), and the goal is to reconstruct spatially masked regions based on the surrounding context. 
These models often take conditionals like class labels \citep{Dhariwal:2021:Diffusion}, text embeddings \citep{Ramesh:2022:CLIP, Nichol:2022:GLIDE}, or segmentation maps \citep{Rombach:2022:LDM}. These conditionals are typically injected via concatenation at the input or embedding injection via FiLM modulation. 
This approach ignores the implicit spatial alignment between the conditionals and the generative target. More recent approaches like ControlNet \citep{Zhang:2023:ControlNet} and BrushNet \citep{Ju:2024:BrushNet} introduce multiscale conditioning mechanisms that are spatially aligned and applied post hoc to pre-trained Stable Diffusion models. 
While these methods preserve spatial alignment, they assume low-dimensional RGB-style inputs with simple spatial masks and do not address the challenges posed by high-dimensional structured data with intricate inter-channel dependencies. Moreover, because their conditioning modules are trained separately from the core generative model, they lack end-to-end coordination between condition encoding and generation. 

\paragraph{Dynamic guidance for conditional diffusion.}
Classifier-free guidance (CFG) \citep{Ho:2021:CFG} introduces a flexible training scheme for conditional diffusion models by randomly dropping conditioning inputs and jointly training the model on both conditional and unconditional objectives. While originally developed for low-dimensional conditioning signals such as class labels, the core idea can be generalized: using input masking during training to enable flexible guidance at test time. We adopt this principle in the form of random-masking guidance, where the spatial conditions are randomly masked during training. Specifically, the model observes a random subset of input channels and is trained to reconstruct the full panel. This dynamic masking encourages the model to learn a unified generative model that generalizes across different conditionals, making it suitable for tasks where the available condition varies across samples. This approach enables multi-channel prediction and supports generation under arbitrary conditioning subsets without retraining or architectural changes.

\paragraph{Channel-aware attention mechanisms.}
While channel-wise attention has been explored in vision architectures like SENet \citep{Hu:2018:SENet}, most diffusion-based models focus on spatial attention and overlook channel-wise relationships.
However, this becomes a significant limitation in multi-channel biological data. A few recent works in diffusion-based colorization have begun to explore this direction: FCNet \citep{Zhang:2023:FCNet} introduces spatially decoupled color representations tailored to facial regions, and ColorPeel \citep{Butt:2024:ColorPeel} learns disentangled geometric shapes and color embeddings in latent space. These models can be thought of as coarse region-level attention over RGB channels and are not designed for tasks involving dozens or hundreds of semantically distinct channels. 

\paragraph{Structured image generation in science.}
Generative models for scientific data, such as multiplexed tissue imaging and spatial transcriptomics, have primarily relied on deterministic architectures, including CNNs and transformers. A few recent methods have extended probabilistic modeling to single-cell multi-omics prediction (e.g., scMM \citep{Minoura:2021:scMM}, UnitedNet \citep{Tand:2023:UnitedNet}) or protein inference from gene expression \citep{MartinGonzalez:2021:MULTIPLAI}, but do not leverage generative diffusion frameworks. The most recent Stem \citep{Zhu:2025:STEM} incorporates diffusion models (specifically DiT) for spatial transcriptomics. However, Stem conditions only on vector embeddings of RGB histology images that break the spatial alignment, and does not account for inter-channel relationships in the generated modalities. Our work builds on these directions by introducing a controllable diffusion-based model tailored to biological profiling data with structured, multi-channel inputs.

\section{Experiments}
We evaluate our model on both single-cell and spatial imaging datasets to test its capacity for conditional multi-channel generation across modalities, tissues, and prediction modes. Our experiments are designed to test (1) prediction performance against state-of-the-art methods, (2) the ability to model inter-channel relationships in multi-channel spatial data, (3) generalization to unseen channel configurations, (4) cross-dataset generalizability, and (5) the contribution of each model component. We report Pearson correlation ($r$) between predicted and ground truth channels unless otherwise specified.~\looseness=-1

\subsection{Single-cell modality prediction}
We first benchmark our model on the task of predicting protein expression from scRNA-seq profiles, a standard benchmark in single-cell multimodal modeling. We use four publicly available datasets: peripheral blood mononuclear cells (PBMC) \cite{hao2021pbmc}, cord blood mononuclear cells (CBMC) \cite{stoeckius2017cbmc}, bone marrow mononuclear cells (BMMC) \cite{Lance:2022:OPSCA} and hematopoietic stem and progenitor cells (HSPCs) \cite{nestorowa2016hspc}. Each dataset contains paired single-cell gene expression (mRNA) and surface protein (ADT) measurements. 

We train the model to predict all protein channels from full gene expression vectors. We compare to a suite of state-of-the-art methods: Kernel Ridge Regression \cite{Lance:2022:OPSCA}, MultiVI \cite{Ashuach:2023:MultiVI}, GLUE \cite{Cao:2022:GLUE}, scMM \cite{Minoura:2021:scMM}, and UnitedNet \cite{Tand:2023:UnitedNet}. Performance is reported as average Pearson correlation between predicted and measured expression levels across proteins ($r_p$) and cells ($r_c$). \Cref{tab:single-cell} shows that our model consistently outperforms all baselines across all four CITE-seq datasets. In particular, our model achieves the highest protein-level correlation $r_p$, which is the more biologically relevant metric, in each setting. We further evaluated the performance of a distilled version of our model capable of performing fast, 1-step inference (via SiD \cite{Zhou:2023:SiD, Zhou:2024:SiDA}). Our method remains competitive with 1-step distillation, suggesting it can be readily deployed for use by biologists with low overhead. 

\begin{table}[]
\caption{ Benchmarking our multimodal diffusion approach against existing modality prediction methods on gene-to-protein prediction task on four (PBMC, CBMC, BMNC and HSPCs) datasets. $r_c$ shows the cell-wise correlation with the ground truth, and $r_p$ shows the protein-level correlation.}
    \label{tab:single-cell}
    \centering
    {\begin{tabular}{c c c  c c  c c  c c}
        \toprule
          \textbf{Method} & \multicolumn{2}{c}{\textbf{PBMC}} & \multicolumn{2}{c}{\textbf{CBMC}} & \multicolumn{2}{c}{\textbf{BMNC}} & \multicolumn{2}{c}{\textbf{HSPC}} \\
        & $r_c$               & $r_p$     
        & $r_c$              & $r_p$ & $r_c$               & $r_c$ & $r_p$               & $r_c$         \\
        \midrule
        KRR \cite{Lance:2022:OPSCA} & \textbf{0.908} & 0.646     
        & 0.863 & 0.006 
        & 0.870 & 0.094
        & 0.820 & 0.059
        \\
        MultiVI \cite{Ashuach:2023:MultiVI}           
        & 0.088  & 0.069         
        & 0.103  & 0.032
        & 0.082  & 0.054
        & 0.045  & 0.035
        \\
        GLUE \cite{Cao:2022:GLUE}                 
        & 0.659  & -0.004
        & 0.623  & 0.054
        & 0.589  & 0.012
        & 0.549  & 0.024
        \\
        UnitedNet \cite{Tand:2023:UnitedNet}        
        & 0.870 & 0.518         
        & 0.377 & 0.628
        & 0.625 & 0.634
        & 0.310 & 0.436
        \\
        scMM \cite{Minoura:2021:scMM}
        & 0.793 & 0.521        
        & 0.736 & 0.517
        & 0.853 & 0.625
        & 0.724 & 0.598
        \\
        \midrule
         Ours (500 steps) 
         & 0.880 & \textbf{0.673} 
         & \textbf{0.962} & \textbf{0.763}
         & \textbf{0.879} & \textbf{0.685}
         & \textbf{0.865} & \textbf{0.647}
         \\
         Ours + SiD (1 step)     & 0.874 & 0.672 
         & \textbf{0.962} & 0.759
         & 0.875 & 0.682
         & \textbf{0.865} & 0.642
         \\
        \bottomrule
    \end{tabular}}
\end{table}

\subsection{Spatial protein imputation}
\label{sec:exp_spatial}
\paragraph{Single channel imputation.} To evaluate the model’s performance in spatial contexts, we apply it to high-dimensional tissue images from Imaging Mass Cytometry (IMC), where each pixel corresponds to spatially co-registered multi-protein expression signals. We evaluate our method on two IMC datasets: a lung cancer cohort \cite{yoffe:etal:2025:lung-spatial-imc-visium} (8 patients) and a breast cancer cohort \cite{jackson2020breast} (6 patients), with 43 and 50 co-registered protein channels, respectively. Images are normalized per channel and tiled into to $64 \times 64$ patches, which is further decreased to $16 \times 16$ by pretrained encoder.
\begin{table}
\caption{\label{tab:imc} Comparison of predictive correlation across different methods in spatial prediction tasks. Our method outperforms the diffusion, single-protein, and linear baselines, and diffusion based ControlNet~\cite{Zhang:2023:ControlNet}, domain-specific models like Stem~\cite{Zhu:2025:STEM} and MULTIPLAI~\cite{MartinGonzalez:2021:MULTIPLAI}.}
\label{tab:imc}
\centering
\begin{tabular}{l c c}
\toprule
\textbf{Method} 						& \textbf{Breast} & \textbf{Lung}\\
\midrule
Most correlated protein  				& 0.489 				 & 0.506 \\
Most spatially correlated protein   				& 0.581 				 & 0.627 \\
\midrule
MULTIPLAI \cite{MartinGonzalez:2021:MULTIPLAI}  & 0.353     	 & 0.349 \\
Stem  \cite{Zhu:2025:STEM}      		& 0.403 				 & 0.475 \\
ControlNet \cite{Zhang:2023:ControlNet} & 0.452				 	 & 0.537 \\
\midrule
Ours (single channel ) 					& \textbf{0.667}		 & \textbf{0.703} \\
Ours (multi channel) 					& 0.596 		 		 & 0.647\\
\bottomrule
\end{tabular}
\end{table}
We compare our method to two simple baselines, single-protein predictor and kernel ridge regression, and structured conditional diffusion models, ControlNet~\citep{Zhang:2023:ControlNet}. 
We also include domain-specific methods, STEM~\cite{Zhu:2025:STEM}  and MULTIPLAI~\cite{MartinGonzalez:2021:MULTIPLAI}. For each image, in the single channel prediction setup, we mask one protein channel at a time as the target and use the rest as the spatial condition.

\begin{figure}[t]
\includegraphics[width=1\linewidth]{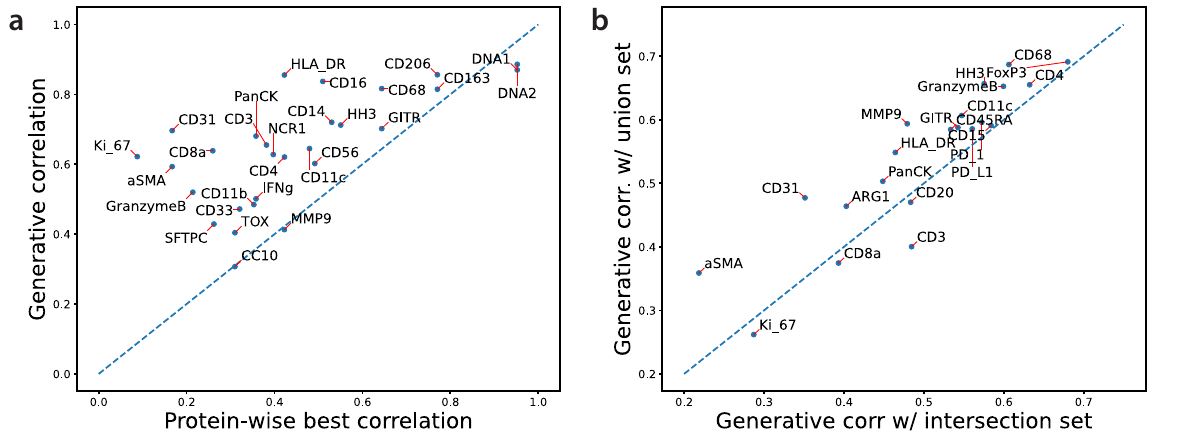} \caption{(a) Benchmarking generative performance (Pearson’s $r$) against the most similar protein on the lung cancer dataset. (b) Cross-dataset generalization study results. Using the union of all protein channels, even those missing from the test dataset, consistently outperforms using only the overlapping set of protein channels.} 
\label{fig:benchmark}
\end{figure}

\Cref{tab:imc} shows our method outperforms all existing approaches on both IMC datasets.
All baseline models fail to outperform the best linear predictor and most fail to outperform the best individual observed protein feature. In the Appendix, we also present hybrid experiments comparing two diffusion models (ControlNet and BrushNet~\citep{Ju:2024:BrushNet}) augmented with elements of our proposed model; results show these methods still do not outperform our full method. This highlights how, unlike in natural RGB images, the spatial structure in biological models presents fundamental challenges to standard diffusion models.
By contrast, our method consistently performs better than the baselines across each individual protein channel, sometimes by more than 300\% (See \cref{fig:benchmark}a for the performance of the multi-channel model).
Fixing the channel known to be missing during training improves performance in the single-protein prediction mode, but the multi-channel prediction model that learns all channels jointly still achieves better performance than any baseline. \Cref{fig:visual_comparison} shows qualitative reconstructions of proteins.
These results validate the effectiveness of random-masking guidance and support practical use cases where multi-channel prediction is required.



\paragraph{Cross-dataset generalization.}
\label{sec:exp_generalization}
We further test our model's ability to learn from datasets with partially overlapped protein channels. Specifically, we combine the lung and breast cancer IMC datasets. The two datasets have $23$ proteins in common, with $18$ unique lung panel proteins and $21$ unique breast panel proteins. We evaluate our model under two training schemes: (1) an intersection setting, where we keep only the proteins observed in both datasets and (2) a union setting, where we include the union of all protein channels and zero-pad missing channels in each dataset. Both setups share the same model architecture and training strategy with random masking, and both predict the intersection set at test time.

As shown in \cref{fig:benchmark}b, the union setup consistently outperforms the intersection counterpart, achieving a higher average Pearson correlation in sample generation. Although increasing the sparsity of supervision, the union setup has a broader coverage and enables the model to learn a richer set of inter-channel dependencies and improves its robustness to missing data. The model effectively leverages partially observed data and maintains performance by training on the union of proteins with zero-padded missing channels. Structurally aligned but partially missing data can still carry informative gradients for joint learning under random-masking guidance, highlighting random-masking guidance as a principled method for multi-dataset integration with empirical advantages.

\subsection{Ablation studies}
We conduct ablation experiments to disentangle the contributions of each architectural component. Using the lung cancer IMC dataset, we compare two feature injection mechanisms: elementwise addition and soft channel-wise attention, as well as three attention configurations: (1) with channel-wise attention only in the UNet blocks, (2) with channel-wise attention only in the output block, and (3) the full model combining both mechanisms.
\begin{table}
\caption{ Abalation studies with 4 different architectual setups on the breast cancer dataset; test score is the avearage over all proteins.}
\label{tab:ablation}
\centering
\baselinestretch 
\begin{tabular}{l c}
\toprule
\textbf{Method} & \textbf{$r$} \\
\midrule
Best Protein Predictor & 0.489 \\
\midrule
Single channel &  0.667 \\
Multi channel & 0.596 \\
\quad w/o output channel attention & 0.581 \\
\quad w/o UNet channel attention & 0.541 \\
\midrule
Condition injection w/ element-wise addition &  0.516\\
Condition injection w/ channel attention & 0.535\\
Base (unconditional)& -0.017 \\
\bottomrule
\end{tabular}
\end{table}

\Cref{tab:ablation} summarizes the contributions of each architectural component. The unconditional baseline performs poorly, with Pearson $r<0.1$. Adding hierarchical feature injection significantly improves performance, and conditional injection via soft channel attention further improves over conventional element-wise addition. Output-space channel attention provides modest gains, while latent-space channel attention leads to more substantial improvements, capturing complex inter-channel dependencies in the latent feature space. These results support integrating both spatially aligned conditioning and channel-wise attention for multi-channel imputation.


\section{Conclusion}
We introduced a diffusion-based generative model for controllable generation of multi-channel biological profiling data, capable of handling arbitrary combinations of observed and missing modalities.
Biological profiling technologies such as IMC and 10X Xenium~\citep{janesick:etal:2023:xenium} are expensive, time-consuming, and require specific panel design, which can only cover a subset of the total molecules of interest at a time. These constraints limit the scientific utility of these technologies. Our controllable generative model offers an \emph{in silico} extension to the pre-designed profiling panel, and has the potential to recover corrupted or incomplete acquisitions. The ability of our model to learn rich priors over distinct tissue types and microenvironments suggests it may make a strong basis for a foundation model for spatial biology. In future work, we plan to explore scaling our model to dozens or hundreds of spatial datasets to enable broad protein channel imputation.


\bibliography{ref}
\bibliographystyle{unsrt}
\end{document}